%% file: c-bandit-arxiv.tex
\documentclass[twocolumn,letterpaper]{article}
\usepackage[utf8]{inputenc}

\pdfoutput=1

\usepackage{arxiv}
\usepackage{comment}
\usepackage{textcomp}
\usepackage{amsmath,mathtools}
\usepackage{amsthm,thmtools,thm-restate}
\usepackage{txfonts}
\usepackage{amssymb} 
\usepackage[scr=rsfso]{mathalfa}
\usepackage{dsfont}
\usepackage{bm}

\usepackage[boxed]{algorithm}
\usepackage{algpseudocode}

\usepackage[inline,shortlabels]{enumitem}
\usepackage{pgfplots}
\usepackage{pgfplotstable}
\usepackage{tikz-cd}
\usepackage{booktabs}

\usepackage[round,colon,authoryear]{natbib}
\usepackage{varioref}
\usepackage[colorlinks=true,citecolor=blue]{hyperref}
\usepackage[capitalise]{cleveref}

\usepackage[disable]{todonotes}

\newcommand{\tinytodo}[2][]{\todo[size=\tiny]{#2}}
\newcommand{\todoy}[2][]{\tinytodo[color=red!20, #1]{Y:\@#2}} 

\newcommand{\Ex}{\mathbb{E}}

\DeclareMathOperator*{\argmax}{arg\,max}
\newcommand\given[1][]{\nonscript\:#1\vert\nonscript\:\mathopen{}\allowbreak}
\DeclarePairedDelimiter\paren\lparen\rparen
\DeclarePairedDelimiter\bracket\lbrack\rbrack
\DeclarePairedDelimiter\abs||
\DeclarePairedDelimiterX\set[1]\{\}{%
  \renewcommand\given{\nonscript\:\delimsize\vert\nonscript\:\mathopen{}\allowbreak}
  #1
}

\algnewcommand{\CommentLine}[1]{\Statex $\triangleright$ #1}

\newcommand{\bset}[1]{\left\{#1\right\}}
\newcommand{\eqn}[1]{\setlength{\abovedisplayskip}{0.15cm}\setlength{\belowdisplayskip}{0.15cm}\begin{align}#1\end{align}}
\newcommand{\eq}[1]{\setlength{\abovedisplayskip}{0.15cm}\setlength{\belowdisplayskip}{0.15cm}\begin{align*}#1\end{align*}}

\newcommand{\ind}[1]{\mathbb{1}\!\!\bset{#1}}

\renewcommand{\P}[1]{\mathbb{P}\left\{#1\right\}}

\newcommand{\iset}[1]{\left[#1 \right]}

\newcommand{\E}[1]{\mathbb{E}\left[ #1 \right]}

\theoremstyle{plain}
\newtheorem{theorem}{Theorem}

\theoremstyle{definition}

\newtheorem{assumption}[theorem]{Assumption}
\newtheorem{remark}[theorem]{Remark}

\theoremstyle{remark}
\theoremstyle{assumption}

\newcommand{\email}[1]{\small\href{mailto:#1}{\nolinkurl{#1}}}

\title{Conservative Optimistic Policy Optimization via Multiple Importance Sampling}

\DeclareRobustCommand{\authorlist}{
  \begin{tabular}[t]{cc}
  Achraf Azize & Othman Gaizi \\
  \email{achraf.azize@polytechnique.edu} &
  \email{othman.gaizi@polytechnique.edu} \\[2ex]
    \multicolumn{2}{c}{\textit{Ecole Polytechnique}}
  \end{tabular}}
\author{\authorlist}

\date{}

\begin{document}

\maketitle{}

\begin{abstract}
    Reinforcement Learning (RL) has been able to solve hard problems such as playing Atari games or solving the game of Go, with an unified approach. Yet modern deep RL approaches are still not widely used in real-world applications. One reason could be the lack of guarantees on the performance of the intermediate executed policies, compared to an existing (already working) baseline policy. In this paper, we propose an online model free algorithm that solves conservative exploration in the policy optimization problem. We show that the regret of the proposed approach is bounded by $\tilde{\mathcal{O}}(\sqrt{T})$ for both discrete and continuous parameter spaces.
\end{abstract}

\section{Introduction}
\label{sec:intro}
\input{introduction}

\section{Preliminaries}
\label{sec:prelim}
\input{preliminaries}

\section{Problem Formalization}
\label{sec:form}
\input{problem_formalization}

\section{Algorithms}
\label{sec:stochastic}
\input{algorithms}

\section{Regret Analysis}
\label{sec:adversarial}
\input{regret_analysis}

\section{Experiments}
\label{sec:experiments}
\input{experiments}

\section{Conclusion}
\input{conclusion}

\renewcommand{\bibsection}{\subsection*{\refname}}

\bibliography{references}
\bibliographystyle{plainnat}

\appendix

\twocolumn[
\thispagestyle{plain}
\centering\textbf{\Large Appendix}
\vspace{2\baselineskip}
]
\input{appendix}

\end{document}

%% file: introduction.tex
The goal of reinforcement learning (Sutton and Barto, 1998)
is to learn optimal policies for sequential decision problems,
by optimizing a cumulative future reward signal. Policy optimization (PO) is a class of RL algorithms that models explicitly the policy (behaviour) of an agent as a parametric mapping from states to actions. This class is usually suited for continuous tasks, where the states and actions are modeled as real numbers.  

While the problem of finding an optimal policy with the least amount of interactions with the environment is very important and has been widely studied in the PO literature, the problem of controlling the performance of the agent during learning is still a challenge. This online policy optimization is extremely relevant when an agent is unable to learn before being deployed to the real world (e.g recommendation system). 

In this online setting,
the agent needs to trade-off exploration and
exploitation while interacting with the environment. The agent is willing
to give up rewards for actions improving his knowledge
of the environment. Therefore, there is no guarantee
on the performance of policies generated by the algorithm, especially in the initial phase where the uncertainty about the environment is maximal. This is a major obstacle that prevents the application of RL in domains where hard constraints
(e.g., on safety or performance) are present. Examples
of such domains are digital marketing, healthcare, finance and robotics. For a vast number of domains, it
is common to have a known and reliable baseline policy
that is potentially suboptimal but satisfactory. Therefore, for applications of RL algorithms, it is important
that are guaranteed to perform at least as well as the
existing baseline.

This setting has been studied in multi-armed bandits (\cite{banditconservative}), and also very well defined in the RL case (\cite{garcelon20}). In this paper, we first summarize algorithmic ideas from \cite{papini19}, formalize the problem of Conservative Policy Optimization, propose algorithms that solve this problem in both discrete and compact parameter space and finally show that those algorithms yield a sublinear regret  $\tilde{\mathcal{O}}(\sqrt{T})$.


%% file: preliminaries.tex
\subsection{The Policy Optimization Problem}
\label{sec:po}

In this section, we will use the same formalisation of policy optimization introduced by \cite{papini19}.

Let $\mathcal{X} \subseteq \mathbb{R}^d$ be the arm set and $(\Omega , \mathcal{F}, P )$ a probability space. Let $\{ \mathit{Z}_{\boldsymbol{x}}: \Omega \rightarrow \mathcal{Z} ~ | ~ x \in \mathcal{X} \}$ be a set of continuous random vectors parametrized by $\mathcal{X}$ , with common sample space
$\mathcal{Z} \subseteq \mathbb{R}^m $. We denote with $p_{\boldsymbol{x}}$ the probability density function of $\mathit{Z}_{\boldsymbol{x}}$. Finally, let $f : \mathcal{Z} \rightarrow \mathbb{R}$ be a bounded payoff
function, and $\mu (\boldsymbol{x}) = \mathbb{E}_{z \sim p_{\boldsymbol{x}} } [f(z)]$ its expectation under
$p_{\boldsymbol{x}}$. For each iteration $t = 0,\dots, T$ , we select an arm $\boldsymbol{x}_t$,
draw a sample $z_t$ from $p_{\boldsymbol{x}_t}$
, and observe payoff $f(z_t)$, up
to horizon H. The goal is to maximize the expected total
payoff:
\begin{equation}
    \max_{\boldsymbol{x}_0,\dots,\boldsymbol{x}_T \in \mathcal{X}} \sum_{t=0}^{T} \mathbb{E}_{z_t \sim p_{\boldsymbol{x}_t} } [f(z_t)] = \max_{\boldsymbol{x}_0,\dots,\boldsymbol{x}_T \in \mathcal{X}} \sum_{t=0}^{T} \mu (\boldsymbol{x}_t)
    \label{eq:1}
\end{equation}

In action-based PO, $\mathcal{X}$ corresponds to the parameter space $\Theta$ of a class of stochastic policies $\left \{ \pi_{\theta} : \theta \in \Theta  \right \}$, $\mathcal{Z}$ to the set $\mathcal{T}$ of possible trajectories ($\tau = \left [ s_0, a_0, s_1, a_1, \dots,  s_{H-1}, a_{H-1} \right ]$), $p_{\boldsymbol{x}}$ to the density $p_{\theta}$ over trajectories induced by policy $\pi_{\theta}$, and $f(z)$ to the cumulated
reward $\mathcal{R}(\tau) = \sum_{h=0}^{H-1} \mathcal{R}(s_h,a_h) = \sum_{h=0}^{H-1} r_{h+1}$. 

In parameter-based PO, $\mathcal{X}$ corresponds to the
hyperparameter space $\Xi$  of a class of stochastic hyperpolicies $\{ \nu_{\xi} : \xi \in \Xi \}$, $\mathcal{Z}$ to the cartesian product $\Theta \times \mathcal{T}$, $p_{\boldsymbol{x}}$
to the joint distribution $p_{\boldsymbol{\xi}} (\boldsymbol{\theta}, \tau) := \nu_{\boldsymbol{\xi}}(\boldsymbol{\theta}) p_{\boldsymbol{\theta}}(\tau)$, and $f(z)$ to the cumulated
reward $\mathcal{R}(\tau)$.

\begin{figure}
    \centering
    \includegraphics[width=1\linewidth]{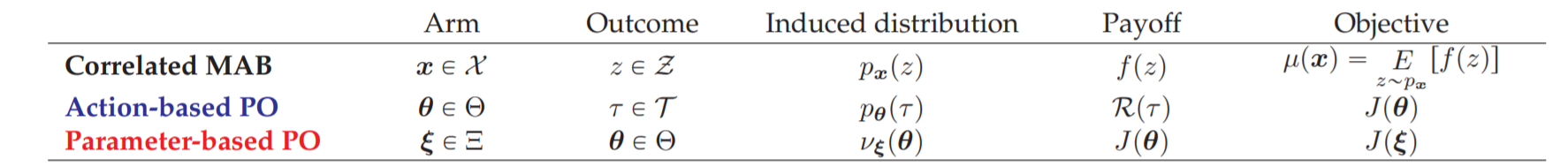}
    \caption{The Policy Optimisation problem formalization}
    \label{fig:mab_form}
\end{figure}
Both cases are summarized in figure \ref{fig:mab_form}. The peculiarity of this framework, compared to the classic
MAB one, is the special structure existing over the arms. In
particular, the expected payoff $\mu$ of different arms is correlated thanks to the stochasticity of $p_{\boldsymbol{x}}$ on a common
sample space $\mathcal{Z}$. In the following, we will use multiple importance sampling to exploit this correlation and
guarantee efficient exploration.

\subsection{Robust Multiple Importance Sampling Estimation}
Importance sampling is a
technique that allows estimating the expectation of a function under some target or proposal distribution with samples
drawn from a different distribution, called behavioral.

Let $\mathit{P}$ and $\mathit{Q}$ be probability measures on a measurable space
$(\mathcal{Z}, \mathcal{F}$, such that $\mathit{P}  \ll  \mathit{Q}$ (i.e., $\mathit{P}$ is absolutely continuous w.r.t. $\mathit{Q}$). Let
$p$ and $q$ be the densities of $\mathit{P}$ and $\mathit{Q}$, respectively, w.r.t. a
reference measure, and the importance weight $\omega_{P / Q} = \frac{p}{q}$. Given a bounded function $f : \mathcal{Z} \rightarrow \mathbb{R}$, and a set of
i.i.d. outcomes $z_1, . . . , z_N$ sampled from $\mathit{Q}$, the importance
sampling estimator of $\mu := \mathbb{E}_{z \sim P} [f(z)]$ is:
\begin{equation}
    \widehat{\mu}_{IS} := \frac{1}{N} \sum_{i = 1}^{N} \omega_{P / Q}(z_i) f(z_i)
    \label{eq:2}
\end{equation}
which is an unbiased estimator, i.e.,
$\mathbb{E}_{z_i \overset{\text{iid}}{\sim} Q} [\widehat{\mu}_{IS}] = \mu$.

Multiple importance sampling is a generalization of the importance sampling technique
which allows samples drawn from several different behavioral distributions to be used for the same estimate. Let $\mathit{Q}_1, \dots, \mathit{Q}_K$ be all probability measures over the same probability space as $\mathit{P}$, and $\mathit{P}  \ll  \mathit{Q}_k$ for $k = 1, \dots, K$. Given $N_k$
i.i.d. samples from each $\mathit{Q}_k$, the Balance Heuristic Multiple Importance Sampling (MIS) estimator is:
\begin{equation}
    \widehat{\mu}_{BH} := \sum_{k = 1}^{K} \sum_{i = 1}^{N_k} \frac{p(z_{ik})}{\sum_{j=1}^{K} N_j q_j(z_{ik})} f(z_{ik})
    \label{eq:3}
\end{equation}
which is also an unbiased estimator of $\mu$.

Recently it has been observed that, in many cases of interest, the plain estimators \ref{eq:2} and \ref{eq:3}
present problematic tail behaviors,
preventing the use of exponential concentration inequalities. A common heuristic to address this problem consists
in truncating the weights :
\begin{equation}
    \Breve{\mu}_{IS} := \frac{1}{N} \sum_{i = 1}^{N} \min \{ M ,\omega_{P / Q}(z_i)\} f(z_i)
    \label{eq:4}
\end{equation}
where $M$ is a threshold to limit the magnitude of the
importance weight. Similarly, for the multiple importance
sampling case, restricting to the BH, we have:
\begin{equation}
    \Breve{\mu}_{BH} := \frac{1}{N} \sum_{k = 1}^{K} \sum_{i = 1}^{N_k} \min \left \{  M, \frac{p(z_{ik})} {\sum_{j=1}^{K} \frac{N_j}{N} q_j(z_{ik})} \right \}  f(z_{ik})
    \label{eq:5}
\end{equation}

Clearly, since we are changing the importance weights, we
introduce a bias term, but, by reducing the range of the
estimate, we get a benefit in terms of variance.

Intuitively, we can allow larger truncation thresholds M
as the number of samples $N$ increases. The results from \cite{papini19}
state that, when using an adaptive threshold depending on
$N$, we are able to reach exponential concentration: with at least probability $1-\delta$, we have:
\[
 \Breve{L}(\delta) \leq \mu \leq \Breve{U}(\delta)
\]
with:
\begin{equation}
    \Breve{U}(\delta) = \Breve{\mu}_{BH} + \left \| f \right \|_{\infty} \left ( \sqrt2 + \frac{4}{3} \right ) \left ( \frac{d_2(P \| \Phi   ) \log(\frac{2}{\delta})}{ N} \right ) ^{\frac{1}{2}}
\end{equation}
and
\begin{equation}
    \Breve{L}(\delta) = \Breve{\mu}_{BH} - \left \| f \right \|_{\infty} \left ( \sqrt2 + \frac{1}{3} \right ) \left ( \frac{d_2(P \| \Phi   ) \log(\frac{2}{\delta})}{ N} \right ) ^{\frac{1}{2}}
\end{equation}
such that:
\begin{align*}
    &d_2(P \| \Phi   ) = \int_{\mathcal{Z}} \left (  \omega_{P / \Phi}\right )^2 d\Phi = \mathbb{V}ar_{z \sim \Phi} \left [  \omega_{P / \Phi}(z) \right ] + 1 \\
    &\Phi = \sum_{j=1}^{K} \frac{N_j}{N} Q_j ~~ \text{is a mixture model}
\end{align*}
and finally $\Breve{\mu}_{BH}$ is the truncated balance
heuristic estimator as defined in \ref{eq:5}, using $N_k$ i.i.d.
samples from each $Q_k$ and $M_N = \left ( \frac{N d_2(P \| \Phi   )}{\log(\frac{1}{\delta})} \right )^\frac{1}{2} $
with $N =  \sum_{j=1}^{K} N_k$


%% file: problem_formalization.tex
In this section, we will present the conservative exploration formalization with respect to The Policy Optimization Problem defined in \ref{sec:po}.
But first, we need to make two assumptions on the baseline arm.

\begin{assumption}
We suppose that the baseline arm $\boldsymbol{x}_b$ is parametrized i.e. $\boldsymbol{x}_b \in \mathcal{X}$ 
\end{assumption}
In action-based PO, this is equivalent to supposing that there exists a parameter $\theta_b$ such that the baseline policy $\pi_b$ is $\pi_b = \pi_{\theta_b}$.\\
In parameter-based PO, this is equivalent to supposing that there exists a parameter $\xi_b$ such that the baseline hyperpolicy $\nu_b$ is $\nu_b = \nu_{\xi_b}$.\\
In both cases, it is always interesting to find a parameter space that contains the baseline.

\begin{assumption}
We will initially assume that the algorithms
know $\mu_b$ the expected reward of the default arm ($\mu_b = \mu(\boldsymbol{x}_b)$). 
\end{assumption}
This is reasonable in situations where the default action has been used for a long time and is well-characterized.

In PO, we will use this form of conservative constraint:
\begin{equation}
    \forall t \in \left \{ 1,\dots, T \right \} ~~ \sum_{i=0}^{t-1} \mu(\boldsymbol{x}_i) \geq (1 - \alpha) t \mu(\boldsymbol{x}_b)
    \label{eq:8}
\end{equation}
for some $\alpha \in [0,1]$\\
and budget:
\begin{equation}
    B_t = \sum_{i=0}^{t-1} \mu(\boldsymbol{x}_i) - (1 - \alpha)  t \mu(\boldsymbol{x}_b)
    \label{eq:budget}
\end{equation}
In action-based PO, this is equivalent to the constraint defined by \cite{garcelon20} in the finite horizon case, since in action-based PO, $\mu(x_i) = J(\theta_i) = V^{\pi_{\theta_i}}(s)$ if we always start from the same initial state s,
or $\mu(x_i) = \mathbb{E}_{s \sim \rho} \left [  V^{\pi_{\theta_i}}(s) \right ]$ if $\rho$ is the distribution of initial states.

The requirement in \ref{eq:8} is often too strict in practice. We could relax the constraint by only verifying the condition \ref{eq:8} at some “checkpoints”. A simple case is where the checkpoints are equally
spaced every $C^h$ steps.
\begin{equation}
    \forall k > 0 ~~ \sum_{i=0}^{kC^h - 1} \mu(\boldsymbol{x}_i) \geq (1 - \alpha) ~  kC^h \mu(\boldsymbol{x}_b)
    \label{eq:10}
\end{equation}
In order to determine whether an action at t is safe at a time $t \in [kC^h,(k + 1)C^h -1]$ of a phase $k \in \mathbb{N}$,
we want to
ensure that by playing the baseline arm until the next checkpoint (i.e., until $(k + 1)C^h -1$) , the algorithm would meet the
condition \ref{eq:10}. Formally, at any step $t$, we replace \ref{eq:10} with:
\begin{equation}
    \forall t > 0 ~~ \sum_{i=0}^{t} \mu(\boldsymbol{x}_i) + \alpha ((k + 1)C^h -1 - t)\mu(\boldsymbol{x}_b) \geq (1 - \alpha) ~  (t+1) \mu(\boldsymbol{x}_b)
    \label{eq:11}
\end{equation}

Our objective is to design algorithms that maximize the cumulative reward \ref{eq:1} while simultaneously satisfying the constraint \ref{eq:8} (or \ref{eq:11}).
The Conservative Policy optimization problem is then:
\begin{equation*}
    \max_{\boldsymbol{x}_0,\dots,\boldsymbol{x}_T \in \mathcal{X}} \sum_{t=0}^{T} \mathbb{E}_{z_t \sim p_{\boldsymbol{x}_t} } [f(z_t)] = \max_{\boldsymbol{x}_0,\dots,\boldsymbol{x}_T \in \mathcal{X}} \sum_{t=0}^{T} \mu (\boldsymbol{x}_t)
    \label{eq:100}
\end{equation*}
such that:
\begin{equation*}
    \forall t \in \left \{ 1,\dots, T \right \} ~~ \sum_{i=0}^{t-1} \mu(\boldsymbol{x}_i) \geq (1 - \alpha) t \mu(\boldsymbol{x}_b)
    \label{eq:80}
\end{equation*}


%% file: algorithms.tex
In this section, we will propose algorithms that use use the mathematical tools presented
in the Preliminaries \ref{sec:prelim} to build an optimistic estimation of $\mu$, choose the most optimistic arm, play that arm if is safe (by checking if a lower bound on the budget \ref{eq:budget} is positif) or otherwise play the baseline arm.

Like discussed in Preliminaries \ref{sec:prelim}, we will use robust multiple importance sampling to capture the correlation among the arms. To simplify the notation, we treat each sample
$\boldsymbol{x}$ as a distinct one and corresponds to the case $K = t - 1$ and $N_k = 1$. Hence,
at each iteration t:
\begin{align}
    \Check{\mu}_{t}(\boldsymbol{x}) = \sum_{k = 0}^{t-1} \min \left \{M_t, \frac{p_{\boldsymbol{x}}(z_{k})} {\sum_{j=0}^{t-1}  p_{\boldsymbol{x}_j}(z_{k})} \right \}  f(z_{k})
\end{align}
where $M_t = \left ( \frac{t d_2(P \| \Phi_t  )}{\log(\frac{2}{\delta_t})} \right )^\frac{1}{2}$ and $ 
\Phi_t = \frac{1}{t} \sum_{k=0}^{t-1} p_{\boldsymbol{x}_k}$ and 
\begin{align}
    &\Check{U}_t(\boldsymbol{x} ,\delta_t) := \Check{\mu}_{t}(\boldsymbol{x}) + \left \| f \right \|_{\infty} \left ( \sqrt2 + \frac{4}{3} \right ) \left ( \frac{d_2(p_{\boldsymbol{x}} \| \Phi_t   ) \log(\frac{2}{\delta_t})}{t} \right ) ^{\frac{1}{2}} \label{eq:upper}\\
    &\Check{L}_t(\boldsymbol{x} ,\delta_t) := \Check{\mu}_{t}(\boldsymbol{x}) - \left \| f \right \|_{\infty} \left ( \sqrt2 + \frac{1}{3} \right ) \left ( \frac{d_2(p_{\boldsymbol{x}} \| \Phi_t   ) \log(\frac{2}{\delta_t})}{t} \right ) ^{\frac{1}{2}}
    \label{eq:lower}
\end{align}
are upper and lower bounds respectively on $\mu(x)$ with probability $1- \delta_t$.

Since $\mu_b$ is supposed known, we take:
\begin{equation}
    \forall t ~~ \Check{U}(\boldsymbol{x}_b ,\delta_t) = \Check{L}(\boldsymbol{x}_b ,\delta_t) = \mu_b
\end{equation}

In this context, in order to check with high probability that the conservative constraint is verified, we check that a lower-bound on the budget is positif. Pseudo-code for this first version is Algorithm \ref{alg:cucb}

\begin{algorithm}
\caption{Conservative OPTIMIST}\label{alg:cucb}
\begin{algorithmic}[1]
\State \textbf{Input: $\boldsymbol{x}_b$, $\mu_b$, $(\delta_t)_{t=1}^{T}$}
\State Draw sample $z_0 \sim p_{\boldsymbol{x}_b}$ and observe $f(z_0)$
\For{$t = 1,\dotsc,T$}
\State Select arm $\boldsymbol{x}_t \in \arg \max _{\boldsymbol{x} \in \mathcal{X} } \Check{U}(\boldsymbol{x} ,\delta_t)$
\State $\Check{B}_t \gets \sum_{i=0}^{t-1} \Check{L}_t(\boldsymbol{x}_i ,\delta_t) + \Check{L}_t(\boldsymbol{x}_t ,\delta_t)  - (1 - \alpha)  (t+1) \mu(\boldsymbol{x}_b)$ 
\If{$\Check{B}_t \geq 0$}
\State Draw sample $z_t \sim p_{\boldsymbol{x}_t}$ and observe $f(z_t)$
\Else
\State Draw sample $z_t \sim p_{\boldsymbol{x}_b}$ and observe $f(z_t)$
\EndIf
\EndFor
\end{algorithmic}
\end{algorithm}

\begin{figure}
    \centering
    \includegraphics[width=1\linewidth]{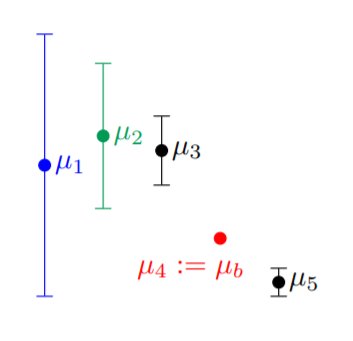}
    \caption{The OPTIMIST arm in blue does not satisfy the conservative condition, thus Alogithm \ref{alg:cucb} will play the basline action. However, arm2 and arm3 are considered "safe arm" . And the Algorithm \ref{alg:cucb3} will choose arm2; which is still conservative, and yields less regret than the baseline (arm chosen by Alogithm \ref{alg:cucb}) } 
    \label{fig:2}
\end{figure}

We can improve the limitation of having a two-step selection strategy by calculating first the set of safe arms (arms that verify that the budget is positif), then take the most optimistic arm from this set. This strategy yields more reward while still being conservative. Figure \ref{fig:2} (taken from \cite{garcelonimproved} ) shows an example to build the intuition for this. The pseudo-code for it is in Algorithm \ref{alg:cucb3}

\begin{algorithm}
\caption{Improved Conservative OPTIMIST}\label{alg:cucb3}
\begin{algorithmic}[1]
\State \textbf{Input: $\boldsymbol{x}_b$, $\mu_b$, $(\delta_t)_{t=1}^{T}$}
\State Draw sample $z_0 \sim p_{\boldsymbol{x}_b}$ and observe $f(z_0)$
\For{$t = 1,\dotsc,T$}
\State $\Check{B}_t(\boldsymbol{x}) = \sum_{i=0}^{t-1} \Check{L}_t(\boldsymbol{x}_i ,\delta_t) + \Check{L}_t(\boldsymbol{x} ,\delta_t)  - (1 - \alpha)  (t+1) \mu(\boldsymbol{x}_b)$
\State $\mathcal{C}_t = \left \{\boldsymbol{x} \in \mathcal{X}: \Check{B}_t(\boldsymbol{x}) \geq 0 \right \}$
\State Select arm $\boldsymbol{x}_t \in \arg \max _{\boldsymbol{x} \in \mathcal{C}_t } \Check{U}(\boldsymbol{x} ,\delta_t)$
\State Draw sample $z_t \sim p_{\boldsymbol{x}_t}$ and observe $f(z_t)$
\EndFor
\end{algorithmic}
\end{algorithm}

The optimization step (line 4 in \ref{alg:cucb} or lines 5 and 6 in \ref{alg:cucb2}) may be very difficult when
X is not discrete as $\Check{U}(\boldsymbol{x} ,\delta_t)$ is non-convex and non-differentiable. Global
optimization methods could be applied at the cost of giving
up theoretical guarantees. In practice, this direction may
be beneficial, but instead, like proposed by \cite{papini19}, we could adapt to the compact case by using a general discretization
method. The key intuition is to make the discretization progressively finer.
The pseudocode for this variant is:
\begin{algorithm}
\caption{Improved Conservative OPTIMIST 2}\label{alg:cucb2}
\begin{algorithmic}[1]
\State \textbf{Input}: $\boldsymbol{x}_b$, $\mu_b$, $(\delta_t)_{t=1}^{T}$, discretization schedule $(\tau_t)_{t=1}^{T}$
\State Draw sample $z_0 \sim p_{\boldsymbol{x}_b}$ and observe $f(z_0)$
\For{$t = 1,\dotsc,T$}
\State $\Check{B}_t(\boldsymbol{x}) = \sum_{i=0}^{t-1} \Check{L}_t(\boldsymbol{x}_i ,\delta_t) + \Check{L}_t(\boldsymbol{x} ,\delta_t)  - (1 - \alpha)  (t+1) \mu(\boldsymbol{x}_b)$
\State Discretize $\mathcal{X}$ with a uniform grid $\tilde{\mathcal{X}}_t$ of $\tau_t^{d} $ points
\State $\tilde{\mathcal{C}_t} = \left \{\boldsymbol{x}  \in \tilde{\mathcal{X}}_t: \Check{B}_t(\boldsymbol{x}) \geq 0 \right \}$
\State Select arm $\boldsymbol{x}_t \in \arg \max _{\boldsymbol{x} \in \tilde{\mathcal{C}_t}} \Check{U}(\boldsymbol{x} ,\delta_t)$
\State Draw sample $z_t \sim p_{\boldsymbol{x}_t}$ and observe $f(z_t)$
\EndFor
\end{algorithmic}
\end{algorithm}

%% file: regret_analysis.tex
Let $Regret(T) = \sum_{i=0}^{T} \Delta_t$
be the total regret with $\Delta_t = \mu(\boldsymbol{x}^{\star}) - \mu(\boldsymbol{x}_t)$ and $\boldsymbol{x}^{\star} \in \arg \max_{\boldsymbol{x} \in \mathcal{X}} \mu(\boldsymbol{x}) $. 

In the following, we will show that Algorithm \ref{alg:cucb} yields
sublinear regret under some mild assumptions (same assumptions as in \cite{papini19}). The proofs
combine techniques from \cite{papini19} and \cite{banditconservative} and are reported in Appendix \label{sec:proof-thm-regret}. First, we need
the following assumption on the Renyi divergence:

\begin{assumption}
We suppose that the 2-Rényi divergence is uniformely bounded:
\begin{equation}
    \sup_{\boldsymbol{x}_0,\dots,\boldsymbol{x}_T \in \mathcal{X}} d_2(p_{\boldsymbol{x}_t} \| \Phi_t) := v_{\epsilon} < \infty
\end{equation}
with $ \Phi_t = \frac{1}{t} \sum_{k=0}^{t-1} p_{\boldsymbol{x}_k}$
\label{ass:1}
\end{assumption}

\subsection{Discrete arm set}
The case of the discrete arm set ($\left | \mathcal{X} \right | = K \in \mathbb{N}$), besides being convenient for the analysis, is also of
practical interest: Even in applications where $\mathcal{X}$ is naturally
continuous (e.g., robotics), the set of solutions that can be
actually tried in practice may sometimes be constrained
to a discrete, reasonably small, set. In this simple setting,
Algorithm \ref{alg:cucb} achieves a regret $\tilde{\mathcal{O}}\left ( \sqrt T \right )$:

\begin{theorem}
With probability $1- \delta$, Algorithm \ref{alg:cucb} with
confidence schedule $\delta_t = \frac{6 \delta}{t^2 \pi^2 K}$, satisfies the following:
\begin{align}
    \forall t \in \left \{ 1,\dots, T+1 \right \} ~~ \sum_{i=0}^{t-1} \mu(\boldsymbol{x}_i) \geq (1 - \alpha) t \mu(\boldsymbol{x}_b) \\
    Regret(T) \leq \Delta_b + 2 \sqrt{LT} +  \frac{\left \| f \right \|_{\infty} \Delta_b}{\alpha \mu_b} + \frac{4KL}{\alpha\mu_b}
\end{align}
with $L = (a+b)^2 v_{\epsilon} [ 2 \log(T) + \log(\frac{\pi^2 K}{3 \delta}) ]$,\\
$a = \left \| f \right \|_{\infty} \left ( \sqrt2 + \frac{4}{3} \right )$ and  $
b = \left \| f \right \|_{\infty} \left ( \sqrt2 + \frac{1}{3} \right )$
\label{thm:regret}
\end{theorem}

This yields a $\tilde{\mathcal{O}}(\sqrt{T})$ regret.

\subsection{Compact arm set}
We consider the more general case of a compact arm set $\mathcal{X}$.
We assume that $\mathcal{X}$ is entirely contained in a box $\left [ -D, D \right ]^d$, with $D\in \mathbb{R}_{+}$. We
also need the following assumption on the expected payoff:

\begin{assumption}
The expected payoff $\mu$ is Lipschitz continuous, i.e., there exists a constant $P > 0$ such that, for every
$\boldsymbol{x}, \boldsymbol{x'} \in \mathcal{X}$ :
\begin{equation}
    \left | \mu(\boldsymbol{x}) - \mu(\boldsymbol{x'})  \right | \leq P \left \| \boldsymbol{x} - \boldsymbol{x'}   \right \|_1
\end{equation}
\label{ass:2}
\end{assumption}

\begin{theorem}
Under Assumptions \ref{ass:1} and \ref{ass:2}, Algorithm \ref{alg:cucb2} with confidence schedule $\delta_t = \frac{6 \delta}{t^2 \pi^2 \left (  1 + \left \lceil t^\frac{1}{2} \right \rceil^d  \right ) }$ and discretization schedule $\tau_t = \left \lceil t^\frac{1}{2} \right \rceil$
guarantees, with probability at least $1-\delta$:
\begin{align}
    \forall t \in \left \{ 1,\dots, T+1 \right \} ~~ \sum_{i=0}^{t-1} \mu(\boldsymbol{x}_i) \geq (1 - \alpha) t \mu(\boldsymbol{x}_b) \\
    Regret(T) \leq \Delta_b  + 2 \sqrt{L'T}  + \frac{\left \| f \right \|_{\infty} \Delta_b}{\alpha \mu_b} + \frac{8L'}{\alpha\mu_b} 
\end{align}
with $L' = (  (a+b) v_{\epsilon}^{\frac{1}{2}}  \left [  \left (  2 + \frac{d}{2} \right )  \log(T) + d \log(2) + \log(\frac{\pi^2}{3 \delta}) \right ]^{\frac{1}{2}}$ \\

$+  PDd  )^{2}$ 
,\\
$a = \left \| f \right \|_{\infty} \left ( \sqrt2 + \frac{4}{3} \right )$ and  $
b = \left \| f \right \|_{\infty} \left ( \sqrt2 + \frac{1}{3} \right )$
\label{thm:disc}
\end{theorem}
Algorithm \ref{alg:cucb2} achieves a regret $\tilde{\mathcal{O}}\left ( d \sqrt T \right )$:
Unfortunately, the time required for
optimization is exponential in arm space dimensionality $d$.


%% file: experiments.tex
\begin{figure}
    \centering
    \includegraphics[width=1\linewidth]{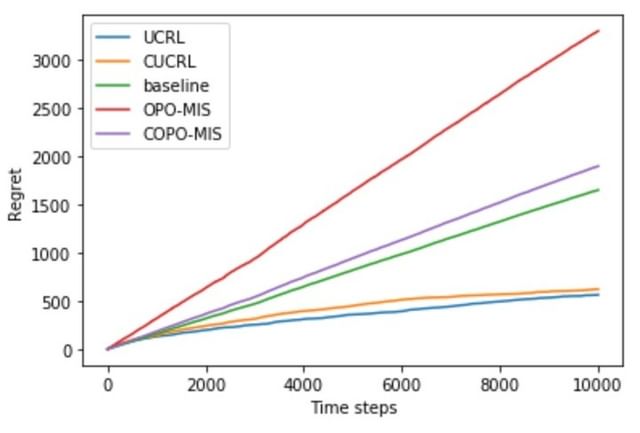}
    \caption{Implementation of OPTIMIST, Conservative OPTIMIST, UCRL, CUCRL on the inventory problem}
    \label{fig:3}
\end{figure}

In this section, we evaluate Algorithm \ref{alg:cucb} compared to CUCRL (\cite{garcelon20}) on the stochastic inventory control problem (\cite{puterman}, Sec. 3.2.1): at the beginning of each month t, a manager has to decide the number of items to order (maximum capacity $M=6$ in order to satisfy a uniform random demand when taking into account ordering and inventory maintenance costs. The optimal policy belongs to the set of thresholds policies characterized by parameters $(\sigma,\Sigma)$ where $\Sigma$ is the target stock and $\sigma$ is the capacity threshold. As a baseline, we decided to take the threshold policy $(\sigma,\Sigma) = (4,4)$, while the optimal one verifies $(\sigma^\star,\Sigma^\star) = (3,6)$. Experiments are run for $T = 10000$ time steps and a conservative level $\alpha = 0.1$, and results displayed in figure \ref{fig:3} are averaged over 20 realizations.

As expected, we can see that the CUCRL algorithm converges a little bit slower than UCRL due to the conservative constraint restricting a free exploration of the environment. However, OPO-MIS (Optimistic Policy Optimization with Multiple Importance Sampling) adapted from \url{https://github.com/WolfLo/optimist} performs really poorly with a linear regret worse than the baseline. This is essentially explained by the authors choice to design linear actor policies in their implementation of the method which is clearly not suitable for this particularly non-linear inventory problem (as the optimal policy is a threshold one). Nonetheless, adding the conservative constraint with COPO-MIS \ref{alg:cucb} (Conservative Optimistic Policy Optimization with Multiple Importance Sampling) enables to not diverge much from the baseline policy and yields a quite similar regret, showing that the conservative constrain is well respected when using \ref{alg:cucb}, and justifying the necessity to add safety constraints when the learning algorithm (OPTIMIST) fails to find a good policy.

%% file: conclusion.tex
In this work, we have studied the problem of conservative exploration in policy optimization using MAB techniques. We
have used algorithmic ideas from \cite{papini19} to propose an online model free algorithm that solves the problem of conservative exploration in RL as defined in \cite{garcelon20}, both the action-based and the parameter-based exploration frameworks, and for both discrete and continuous
parameter spaces. We have proved sublinear regret bounds
for Conservative OPTIMIST under assumptions that are easily met in
practice. The empirical evaluation on the inventory problem showed that the proposed algorithm respect effectively the conservative constraint. However, since the parametrization of the policies (linear policies) in the implementation of \cite{papini19} was not adapted to the inventory case, OPTIMIST failed to reach a good policy. Future work should focus on finding more
efficient parametrization of the policies in the code implementation, but also ways to perform effectively optimization
in the infinite-arm setting.

%% file: appendix.tex
\section{Proof of Theorem~\ref{thm:regret}}
\label{sec:proof-thm-regret}

\begin{proof}

With probability $1- \delta_t$ :
\begin{equation}
    \mu(\boldsymbol{x}) \in \left [\Check{L}_t(\boldsymbol{x} ,\delta_t) ,\Check{U}_t(\boldsymbol{x} ,\delta_t) \right ] ~~ \forall \boldsymbol{x} \in \mathcal{X} ~ \forall t \in \left \{ 1,\dots,T \right \}
    \label{eq:15}
\end{equation}

To ease notation, we will rewrite \ref{eq:15} as:
\begin{equation*}
    \Check{\mu}_{t}(\boldsymbol{x}) - \mu(\boldsymbol{x}) \in \left [-a\beta_t(\boldsymbol{x}, \delta_t), b\beta_t(\boldsymbol{x}, \delta_t) \right ] ~~ \forall \boldsymbol{x} \in \mathcal{X} ~ \forall t \in \left \{ 1,\dots,T \right \} 
\end{equation*}
with $a := \left \| f \right \|_{\infty} \left ( \sqrt2 + \frac{4}{3} \right )$ and  $
b := \left \| f \right \|_{\infty} \left ( \sqrt2 + \frac{1}{3} \right )$ and $
\beta_t(\boldsymbol{x}, \delta_t) := \left ( \frac{d_2(p_{\boldsymbol{x}} \| \Phi_t   ) \log(\frac{2}{\delta_t})}{t} \right ) ^{\frac{1}{2}}$

We take:\\
$F = \bigcap_{k=1}^{K} \bigcap_{t=1}^{T} \left \{   \Check{\mu}_{t}(\boldsymbol{x}_k) - \mu(\boldsymbol{x}_k) \in \left [-a\beta_t(\boldsymbol{x}, \delta_t), b\beta_t(\boldsymbol{x}, \delta_t) \right ] \right \} $

With $\delta_t = \frac{6 \delta}{t^2 \pi^2 K}$, we have:  $\mathbb{P}( F)$
\begin{align*}
    &= 1 -  \mathbb{P}\left (  \bigcup_{k=1}^{K} \bigcup_{t=1}^{T} \left \{   \Check{\mu}_{t}(\boldsymbol{x}_k) - \mu(\boldsymbol{x}_k) \in \left [-a\beta_t(\boldsymbol{x}, \delta_t), b\beta_t(\boldsymbol{x}, \delta_t) \right ] \right \} \right  ) \\
    &\geq 1 - K \sum_{t=1}^{T} \delta_t \\
    &\geq 1 - K \sum_{t=1}^{T} \frac{6 \delta}{t^2 \pi^2 K}\\
    &\geq 1 - \delta
\end{align*}
since $\sum_{t=1}^{T} \frac{1}{t^2} \leq \sum_{t=1}^{\infty} \frac{1}{t^2} = \frac{\pi^2}{6}$.

If $x_t$ is an arm chosen by OPTIMIST, with probability $1-\delta$ we have:  :
\begin{align}
    &\Delta_t = \mu(\boldsymbol{x}^{\star}) - \mu(\boldsymbol{x}_t) \leq \Check{\mu}_{t}(\boldsymbol{x}^{\star}) + a\beta_t(\boldsymbol{x}^{\star}, \delta_t) - \mu(\boldsymbol{x}_t)\\
&\leq \Check{\mu}_{t}(\boldsymbol{x}_t) + a\beta_t(\boldsymbol{x}_t, \delta_t) - \mu(\boldsymbol{x}_t) \\
&\leq (a+b) \beta_t(\boldsymbol{x}_t, \delta_t) = (a+b) \left ( \frac{d_2(p_{\boldsymbol{x}} \| \Phi_t   ) \log(\frac{2}{\delta_t})}{t} \right ) ^{\frac{1}{2}}\\
&\leq (a+b) v_{\epsilon}^{\frac{1}{2}} \left ( \frac{2\log(t) + \log(\frac{\pi^2 K}{3\delta}) }{t} \right ) ^{\frac{1}{2}}\\
&\leq (a+b) v_{\epsilon}^{\frac{1}{2}} \left ( \frac{2\log(T) + \log(\frac{\pi^2 K}{3\delta}) }{t} \right ) ^{\frac{1}{2}}\\
&\leq \sqrt{\frac{L}{t}} \label{eq:dif}
\end{align}
with $L= (a+b)^2 v_{\epsilon} [ 2 \log(T) + \log(\frac{\pi^2 K}{3 \delta}) ] $

Now, we will try to find a bound on the number of times a sub-optimal arm $k$ has been chosen till T: $T_k(T) = \sum_{i=0}^{t} \mathbb{1}_{x_t = x_k}$.
If the arm is never chosen, $T_k(T) = 0$
If it is at least chosen once, let $t$ be the last time $x_k$ has been chosen ($x_k=x_t$), by \ref{eq:dif} we have:
\begin{equation}
    T_k(T) \leq t \leq \frac{L}{{\Delta_t}^2} = \frac{L}{{\Delta_k}^2}
    \label{eq:t}
\end{equation}

The regret can be written as:
\begin{align*}
    &Regret(T) = \sum_{t=0}^{T} \Delta_t \\
    &=  \sum_{{t=1}, t \in \mathcal{S}_T}^{T} \Delta_t ~ + T_b(T) \Delta_b
\end{align*}
with $\mathcal{S}_T = \left \{ t \in \left \{ 1, \dots, T \right \} : x_t \neq x_b  \right \} $ the set of times where the OPTIMIST action was chosen (at t=0 we always choose $x_b$),  
$T_k(T) = \sum_{t=0}^{T} \mathbb{1}_{x_t = x_k}$ the number of times the arm k was chosen till time $T$, and 
$T_b(T) = \sum_{t=0}^{T} \mathbb{1}_{x_t = x_b}$ the number of time the baseline arm was chosen till times $T$.

With probability $1- \delta$, we have:
\begin{align*}
    &\sum_{{t=1}, t \in \mathcal{S}_T}^{T} \Delta_t\\
    &\leq \sum_{{t=1}, t \in \mathcal{S}_T}^{T} \sqrt{\frac{L}{t}}  \\
    &\leq 2 \sqrt{L} \sqrt T
\end{align*}

If $\Delta_b = 0$ then the theorem holds trivially; we therefore
  assume that $\Delta_b > 0$ and find an upper bound for $T_b(T)$.

  Let $\tau = \max \left \{ t\leq T  ~|~  x_t= x_b    \right \}$ be the last round in which
  the default arm is played.  Since $F$ holds and
  $\Check{U}(\boldsymbol{x}_b ,\delta_t)  = \mu_b < \mu^* < \max_i\theta_i(t)$, it follows that
  $x_b$ is never the OPTIMIST choice and the default arm was only played
  because $\Check{B}_\tau < 0$:
  \begin{equation}
      \sum_{k=1, \boldsymbol{x}_k \neq \boldsymbol{x}_b}^K T_k(\tau)\Check{L}(\boldsymbol{x}_k ,\delta_t)   + T_b(\tau - 1)\mu(\boldsymbol{x}_b )  - (1-\alpha) (\tau + 1) \mu(\boldsymbol{x}_b ) < 0
      \label{eq:30}
   \end{equation}   
   we replace $\tau = \sum_{k=1, \boldsymbol{x}_k \neq \boldsymbol{x}_b}^K T_k(\tau) + T_b(\tau - 1)$ in this inequality and rearrange, we have then:
\begin{align*}
    &\alpha T_b(\tau - 1) \mu(\boldsymbol{x}_b ) \\
    & < (1-\alpha) \mu(\boldsymbol{x}_b ) + \sum_{k=1, \boldsymbol{x}_k \neq \boldsymbol{x}_b}^K T_k(\tau) \left ( \left ( 1 - \alpha \right ) \mu(\boldsymbol{x}_b ) - \Check{L}(\boldsymbol{x}_k ,\delta_t) \right ) \\
    &\leq (1-\alpha) \mu(\boldsymbol{x}_b ) + \\
    &\sum_{k=1, \boldsymbol{x}_k \neq \boldsymbol{x}_b}^K T_k(\tau) \left ( \left ( 1 - \alpha \right ) \mu(\boldsymbol{x}_b ) - \mu_k + (a+b) \beta_\tau(\boldsymbol{x}_k, \delta_\tau) \right ) \\
    & \leq \left \| f \right \|_{\infty} + \sum_{k=1, \boldsymbol{x}_k \neq \boldsymbol{x}_b}^K T_k(\tau) \left ( \left ( 1 - \alpha \right ) \mu(\boldsymbol{x}_b ) - \mu_k +    \sqrt{\frac{L}{T_k(\tau)}} \right )\\
    &\leq \left \| f \right \|_{\infty} + \sum_{k=1, \boldsymbol{x}_k \neq \boldsymbol{x}_b}^K a_k T_k(\tau)   + \sqrt{L T_k(\tau)} \\
    &\leq \left \| f \right \|_{\infty} + \sum_{k=1, \boldsymbol{x}_k \neq \boldsymbol{x}_b}^K S_k
    \end{align*}
with $S_k = a_k T_k(\tau)   + \sqrt{L T_k(\tau)} $ and $a_k = \left ( 1 - \alpha \right ) \mu(\boldsymbol{x}_b ) -  \mu_k$ to ease notation.

We have two cases: $a_k > 0$ or $a_k \leq 0$.

If $a_k > 0$: then $\mu_k < \left ( 1 - \alpha \right ) \mu(\boldsymbol{x}_b )$ so arm $k$ is suboptimal and by \ref{eq:t}, we have: $T_k(\tau) \leq T_k(T) \leq \frac{L}{{\Delta_k}^2}$ so then: \\
$S_k \leq  a_k \frac{L}{{\Delta_k}^2} + \frac{L}{{\Delta_k}} \leq \frac{2L}{{\Delta_k}}$ (because $a_k \leq \Delta_k$).

If $a_k \leq 0$: $S_k = a_k T_k(\tau)   + \sqrt{L T_k(\tau)} \leq - \frac{L}{4 a_k}  = \frac{L}{4 (\Delta_b + \alpha \mu_b - \Delta_k) }$ by using $ax^2 + bx \leq - \frac{b^2}{4a}$ when $a\leq 0$.

We can combine both by taking:
\begin{equation*}
    S_k \leq \frac{2L}{{\max (\Delta_k, \Delta_b  - \Delta_k)}} \leq \frac{4L}{\Delta_b}
\end{equation*}
($\mu_b \geq 0 $ and $\max (\Delta_k, \Delta_b  - \Delta_k) \geq \frac{\Delta_b}{2} $).

Finally, we have:
\begin{align*}
    &T_b(T) =  T_b(\tau - 1) + 1 \\
    &\leq 1 +  \frac{\left \| f \right \|_{\infty}}{\alpha \mu_b} +  \frac{4KL}{\alpha \Delta_b \mu_b}
\end{align*}

To conclude, the regret can be bounded with probability $1-\delta$ by:
\begin{align*}
    Regret(T) \leq  2 \sqrt{LT} + \Delta_b + \frac{\left \| f \right \|_{\infty} \Delta_b}{\alpha \mu_b} + \frac{4KL}{\alpha\mu_b}
\end{align*}
\end{proof}

\section{Proof of Theorem~\ref{thm:disc}}
\begin{proof}
The only difference in this case is that \ref{eq:dif} becomes:
\begin{equation}
    \Delta_t \leq (a+b) \beta_t(\boldsymbol{x}_t, \delta_t) + \frac{PDd}{\tau_t}
\end{equation}
(eq (64) from \cite{papini19}, with $P$ the Lipshitz constant)

By replacing $\tau_t = \left \lceil t^\frac{1}{2} \right \rceil$ and $\delta_t = \frac{6 \delta}{t^2 \pi^2 \left (  1 + \left \lceil t^\frac{1}{2} \right \rceil^d  \right ) }$ , we have:
\begin{align*}
    \Delta_t &\leq \frac{(a+b) v_{\epsilon}^{\frac{1}{2}}  \left [  \left (  2 + \frac{d}{2} \right )  \log(T) + d \log(2) + \log(\frac{\pi^2}{3 \delta}) \right ]^{\frac{1}{2}}}{\sqrt{t}} +  \frac{PDd}{\sqrt t}\\
    &= \sqrt{\frac{L'}{t}}
\end{align*}

with
\begin{equation}
    L' = \left (  (a+b) v_{\epsilon}^{\frac{1}{2}}  \left [  \left (  2 + \frac{d}{2} \right )  \log(T) + d \log(2) + \log(\frac{\pi^2}{3 \delta}) \right ]^{\frac{1}{2}}
     +  PDd  \right )^{2}
\end{equation}

In order to reuse the steps from the proof of Theorem \ref{thm:regret}, we need an independent discretization that does not depend on t (otherwise the number of arm at the end is exponential in T which breaks the regret).

To simplify the analyse, we will break the compact arm $\mathcal{X}$ into two regions (or 'big' arms): A first one where $\left \{ \boldsymbol{x} \in \mathcal{X} : \mu(\boldsymbol{x}^\star) - \mu(\boldsymbol{x}) \leq \epsilon  \right \}$ (for a fixed $\epsilon>0$ chosen such that \ref{eq:30} could be exactly split into two terms), that contains the best arm, and a second region (suboptimal 'big' arm) which is just $\left \{ \boldsymbol{x} \in \mathcal{X} : \mu(\boldsymbol{x}^\star) - \mu(\boldsymbol{x}) > \epsilon  \right \}$

Now, we can reususe the same steps from the previous proof, with two 'big' arms, one optimal, and the other suboptimal. The same steps work exactly, with the only difference is having the new $L'$ rather than $L$ and $K=2$.

The regret is then:
\begin{equation}
Regret(T) \leq \Delta_b  + 2 \sqrt{L'T}  + \frac{\left \| f \right \|_{\infty} \Delta_b}{\alpha \mu_b} + \frac{8L'}{\alpha\mu_b}
\end{equation}
\end{proof}
